%% file: manuscript.tex
\documentclass[letterpaper]{article}

\usepackage{amsmath}
\usepackage{natbib,alifeconf}  
\usepackage{url,hyperref,cleveref}
\usepackage{booktabs}
\usepackage{wrapfig}
\usepackage{float}
\usepackage{placeins}
\usepackage{multirow}
\usepackage {threeparttable}

\input{math_commands.tex}
\title{Homeostatic Coupling for Prosocial Behavior}

\author{
    Naoto Yoshida$^{1}$\and
    Kingson Man$^{2}$\\
    \mbox{}\\
    $^1$Kyoto University, Japan \\
    $^2$Feeling Machines LLC, USA\\
    yoshida.naoto.8x@kyoto-u.ac.jp
} 

\begin{document}

\maketitle

\begin{abstract}
When regarding the suffering of others, we often experience personal distress and feel compelled to help\footnote{Preprint. Under review.}. Inspired by living systems, we investigate the emergence of prosocial behavior among autonomous agents that are motivated by homeostatic self-regulation. We perform multi-agent reinforcement learning, treating each agent as a vulnerable homeostat charged with maintaining its own well-being. We introduce an empathy-like mechanism to share homeostatic states between agents: an agent can either \emph{observe} their partner’s internal state ({\bf cognitive empathy}) or the agent’s internal state can be \emph{directly coupled} to that of their partner ({\bf affective empathy}). In three simple multi-agent environments, we show that prosocial behavior arises only under homeostatic coupling – when the distress of a partner can affect one’s own well-being.
Additionally, we show that empathy can be learned: agents can ``decode" their partner's external emotive states to infer the partner's internal homeostatic states. Assuming some level of physiological similarity, agents reference their own emotion-generation functions to invert the mapping from outward display to internal state. Overall, we demonstrate the emergence of prosocial behavior when homeostatic agents learn to ``read" the emotions of others and then to empathize, or feel as they feel.
\end{abstract}
\vspace{-2mm}

Submission type: \textbf{Full Paper}\\

\vspace{-4mm}
Code available at: \url{https://anonymous.4open.science/r/empathy_homeostatic_rl-7E8F}

\section{Introduction}
 
For humans and other social animals, it is often distressing to regard the suffering of others. We feel empathy, sharing in the feelings of others rapidly and automatically through emotional contagion \citep{hsee1993emotional}. Such feelings can provide a strong motivation to reduce the suffering of others, even if it comes at a cost to the self. It has been proposed that tying one’s own welfare to the welfare of others can form the basis of prosocial behavior \citep{de2008putting}. As artificial systems gain in abilities and integrate further into human society, it becomes more urgent to understand the natural mechanisms that implement empathy and which lead to the emergence of altruistic behavior, and to model them in artificial systems. 

\begin{figure}[t]
  \centering
  \includegraphics[width=0.8\linewidth]{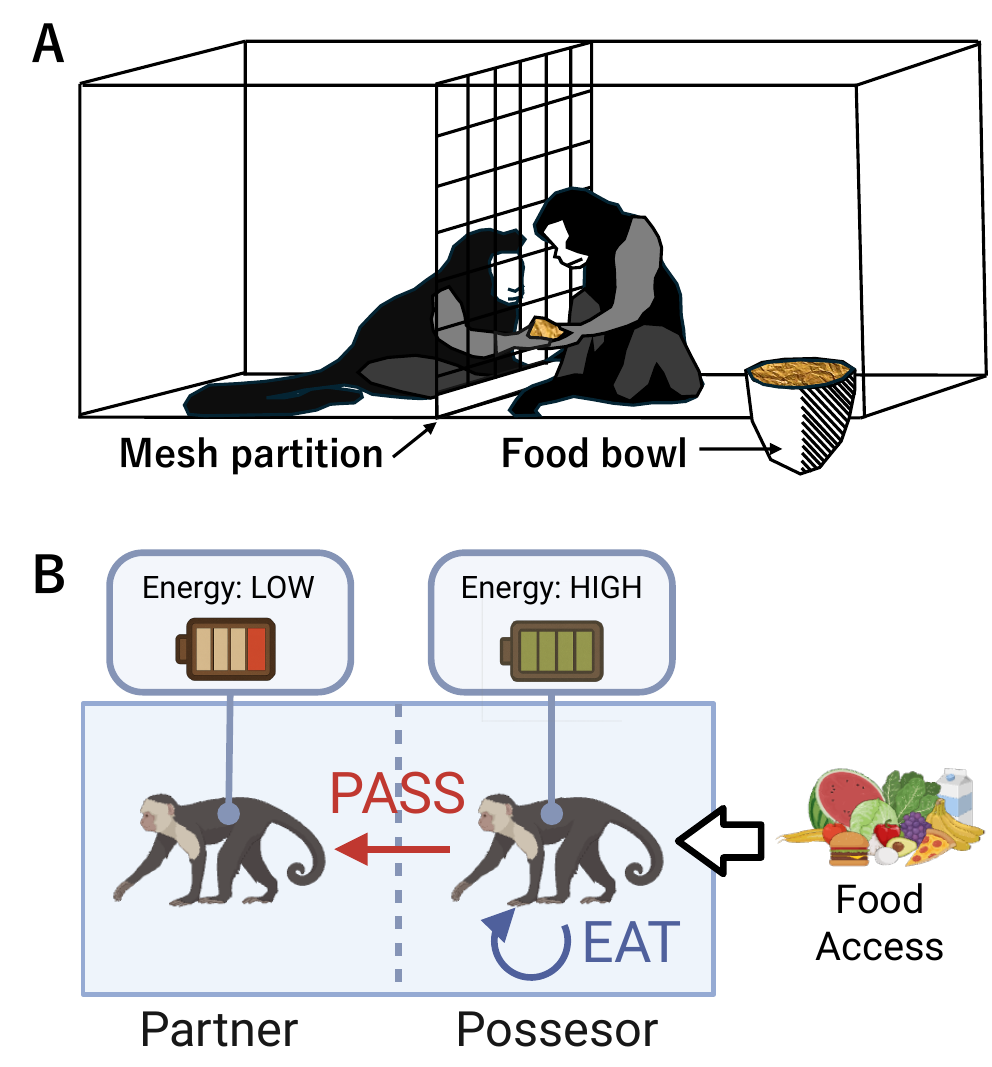}
   \vspace{-3mm}
 \caption{{\bf A}: An experimental setup used in behavioral experiments to test the altruism of monkeys (illustration created based on \citep{de1997food}). {\bf B}: A minimal HRRL environment inspired by behavioral experiments for studying prosocial behavior (food-sharing).}
  \label{fig:foodshare_env}
     \vspace{-5mm}
\end{figure}

Emotions and feelings, whether self- or other-directed, are theorized to arise from homeostasis, the regulation of internal body states within a range compatible with life \citep{damasio1999feeling}. Self-regulatory mechanisms have previously been implemented as a source of external motivation \citep{oudeyer2007typology,baldassarre2011intrinsic}. Here we regard homeostasis \citep{cannon1939wisdom} as an intrinsic and obligatory motivation of all living creatures. Homeostatic-like processes have recently been implemented in reinforcement learning (RL) agents and have resulted in the emergence of integrated behaviors \citep{yoshida2017homeostatic,yoshida2023homeostatic,yoshida2024emergence,yoshida2024synthesising}. Here we start from minimal scenarios of prosocial behavior, originally arising from the field of animal behavior. 

We model social behavior using multi-agent environments, where each agent is formulated as a {\bf homeostat} \citep{ashby1952design,seth2014cybernetic,man2019homeostasis}. We begin with the analysis of a multi-agent toy model inspired by behavioral experiments on food sharing in monkeys (Figure \ref{fig:foodshare_env}). We next propose some requirements for prosocial behavior. In several simulations, we compare the effects of different implementations of empathy \citep{christov2023preventing}, including cognitive empathy, in which an agent can observe the internal state of a partner, and affective empathy, in which an agent’s own internal states are coupled to their partner’s internal states. 

We contribute the following findings: 1) Even in a very simple system, prosocial behavior was not acquired when each agent aimed for homeostasis only within its own body. 2) Prosocial behavior also did not emerge when an agent directly observed the internal states of other agents (cognitive empathy). 3) Prosocial behavior was observed when the agent’s internal state was directly coupled to the internal states of other agents (affective empathy). 
4) Prosocial behavior was acquired as agents learned to better decode the homeostatic states of partner agents from the partner's outward expressions. Our results suggest that, for homeostats motivated by self-regulation, it is necessary to incorporate an additional homeostatic coupling mechanism for prosocial behavior to arise. 

\vspace{-2mm}
\section{Homeostatic Reinforcement Learning}
     \vspace{-1mm}
RL provides a framework to learn behavior in dynamic environments that maximizes the sum of future rewards emitted from the environment \citep{sutton2018reinforcement}. The objective of RL is to obtain a policy $\pi: {\cal S}\rightarrow {\cal A}$ that maximizes the expected value of the weighted cumulative sum of future rewards $\sum_{t=0}^\infty \gamma^t r_t$ for all states $s \in {\cal S}$, based on the experience of the interactions with an environment. Here, $t \in {\mathbb Z}$ is the time step, $r$ is the reward signal, ${\cal S}$ is the set of states in the environment, ${\cal A}$ is the set of actions of the agent.  $0\leq \gamma < 1$ is a positive constant called the discount factor. 

Homeostatic RL \citep{keramati2011reinforcement,keramati2014homeostatic,hulme2019neurocomputational} integrates principles from physiological homeostasis by defining reward as the internally perceived reduction of deviations from homeostatic {\bf set points} \citep{keramati2017cocaine,juechems2019does,uchida2022computational,duriez2023homeostatic}. Concretely, the reward is defined as a quantity proportional to the sequential difference of the {\it drive} $D$, as 
\begin{eqnarray}
r_{t+1} = \beta (D_t - D_{t+1}),
\end{eqnarray}
where $\beta>0$ is the scaling constant \citep{keramati2011reinforcement}. The drive function $D (s^i)$ returns a value greater than or equal to zero, such as the distance between $s^i$ and $s^*$. Here $s^i$ is {\bf interoception} that monitors the internal state of the agent's body \citep{sherrington1906integrative,craig2002you} and $s^*$ is the set point of the interoception. Homeostatic parameters are not arbitrarily defined, but are fundamental to the viability and functionality of the agent \citep{keramati2014homeostatic,man2019homeostasis}. In our conception, homeostatic RL asserts that the agent has a {\bf vulnerable} body. Vulnerability is defined as the circular causality by which homeostatic states can affect the agent’s ability to regulate those states (i.e., it gets harder to take care of oneself as one falls apart). 

\begin{figure}[t]
  \centering
  \includegraphics[width=\linewidth]{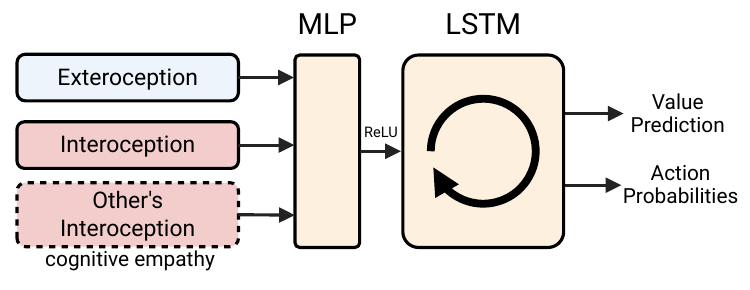}
     \vspace{-5mm}
\caption{Basic network architecture of the agent in this study. Observations include information about the external environment (Exteroception), the agent's own internal conditions (Self-Interoception), and, in the cognitive empathy condition (Other-Interoception), the partner agent's internal conditions.} 
  \label{fig:model}
\end{figure}

\subsection{Agent Architectures in Experiments}
\vspace{-1mm}
In all of our computer experiments, we used Proximal Policy Optimization (PPO, \citep{Schulmanetal_ICLR2016,schulman2017proximal}) as the agent's optimizer. 
The agent's policy model consisted of an encoder of inputs using a multi-layer perceptron and a recurrent connection using LSTM \citep{hochreiter1997long} in all experiments (Figure \ref{fig:model}), and the action selection probability, $\pi$, was calculated by applying the softmax function to the affine transformation from the hidden state.
Value prediction $V_\pi$ is calculated as a scalar output by applying an affine transformation to the hidden state of the LSTM shared with the policy network. 

The models of the agents used in this study all had the same network architecture and optimization was performed with PPO. Agents have {\bf interoception} for their own energy state (explained later) as well as for observations from the outside world (exteroception). In the {\bf cognitive} condition (explained in the later section) of the agent, it also receives interoception from other agents as well. These observations are combined and input to the hidden layer with a linear mapping. The hidden layer takes the ReLU nonlinearity as the activation function and uses it as input to the LSTM. A linear mapping from the output of the LSTM produces value predictions and categorical action selection probabilities using a softmax function. The table shows the network and PPO hyper--parameters for each experiment (Table \ref{tb:fs}--\ref{tb:2d} in Appendix A). 

For the continuous 2D field experiment in the next section, we followed techniques from previous studies on multi-agent systems \citep{foerster2016learning}, optimizing the network using shared weights and experiences across all agents to facilitate learning.

\section{Experiments}

\subsection{Motivational Example: Food Sharing}
It has been reported that brown capuchin monkeys that are separated by a mesh will choose to share food when only one monkey has access to the food \citep{de1997food,de2000attitudinal}. Inspired by these animal behavior experiments on the altruism of monkeys (Figure \ref{fig:foodshare_env}A) \citep{de1997food}, we first created a minimal system for studying prosocial behavior, named the food-sharing environment.  Using this simple system, our computational study explores the minimum configuration in which such sharing behavior occurs in autonomous agents.  

\vspace{-2mm}
\subsubsection{Food Sharing Environment}
An overview of this environment is shown in Figure \ref{fig:foodshare_env}B. We assume two agents: i) a passive agent called the `Partner', corresponding to the monkey in the left side of the cage with no direct access to food, and ii) the active `Possessor', corresponding to the monkey on the right of Figure \ref{fig:foodshare_env}B, and who has access to food. 
Each agent has a binary energy state, High or Low. When the state is High, it changes to Low with probability $p=0.1$ at each time step. If the energy state is Low, it remains Low until an agent eats food and transitions back to High energy. The environment is terminated and reset if any agent remains in the Low energy state for ten consecutive time steps. Energy states are randomly set upon environment initialization. The Possessor has two actions, EAT and PASS. (The Partner has no actions, and only transitions between energy states.) When the Possessor EATs, their energy state is set to High. When the Possessor PASSes, the Partner's energy state is set to High. 

In this environment, only the action optimization of the Possessor is possible. The drive for the homeostasis of the Possessor is $D_{\rm possessor}=-\ln P^*(s^i_t)$. Here, $s^i_t \in \{{\rm High}, {\rm Low} \}$ represents the agent's interoception at time $t$, and $P^*(\cdot)$ is a probability distribution representing the desirability of each state, with $P^*({\rm High})=0.95$ and $P^*({\rm Low})=0.05$. Therefore, the Possessor aims for homeostasis, preferring $s^i={\rm High}$ over $s^i={\rm Low}$.

\vspace{-2mm}
\subsubsection{Analysis of the Environment}
\begin{figure}[t]
  \centering
  \includegraphics[width=0.8\linewidth,bb=0 0 624 512]{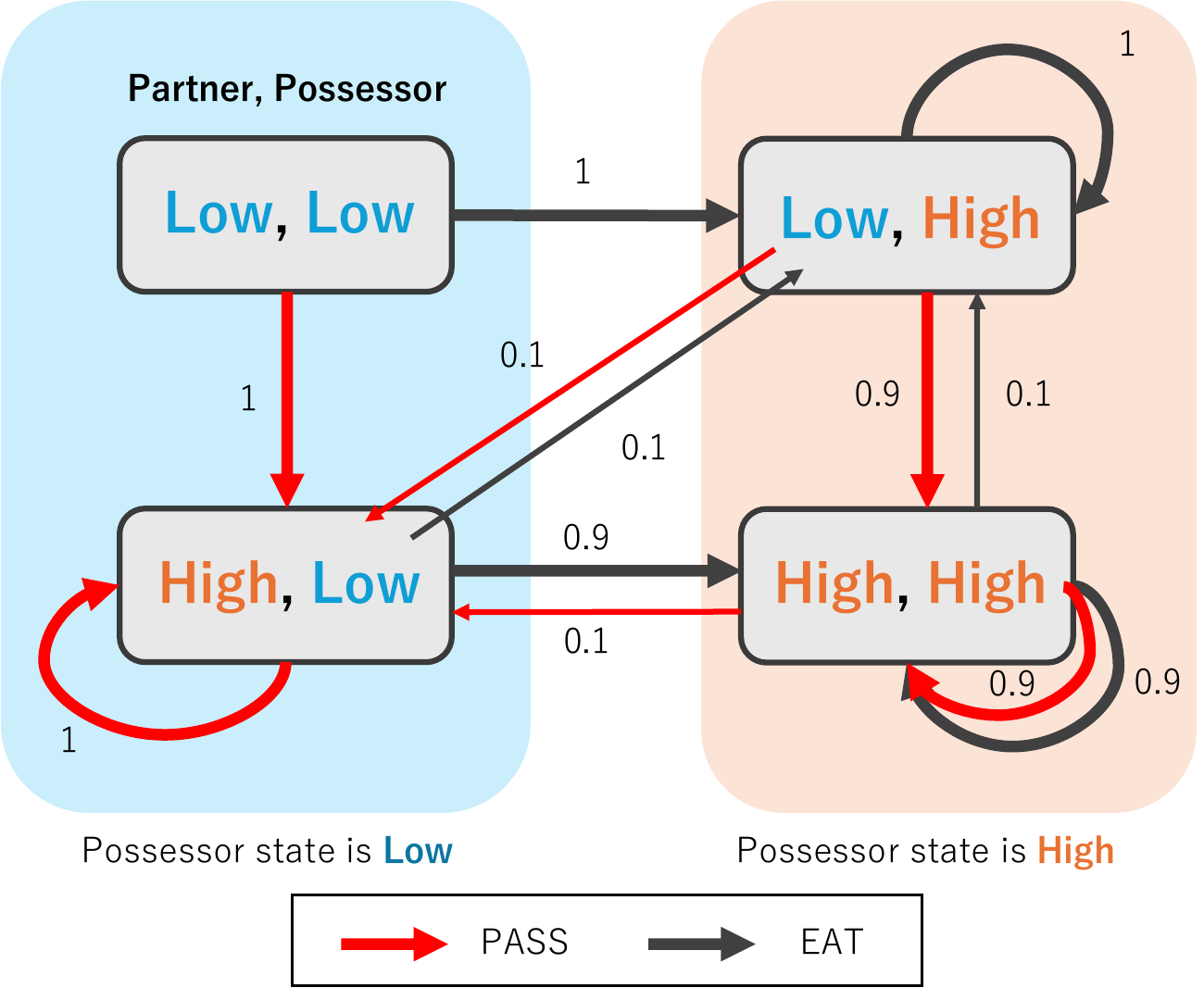}
  \caption{
  State transition diagram of the food sharing environment. Numbers next to arrows are transition probabilities.
  }
     \vspace{-3mm}
\label{fig:afs}
\end{figure}

A simple analysis of the environmental dynamics suggests that prosocial sharing behavior will not emerge if the Possessor is motivated only by its own homeostasis. All the state transitions in the Food Sharing environment is illustrated in Figure \ref{fig:afs}. From this figure, we can see that there is always a risk of the internal state transitioning from High to Low when the Possessor chooses the PASS action. This can be seen from the transitions of the blue and red macro states on the left and right (corresponding to the Possessor's interoception) when the PASS action is chosen. Therefore, it is always optimal for the homeostasis of the Possessor alone to choose the EAT action, and it is suggested that in such a situation, no action to help the Partner will emerge regardless of the probability $p$.

\begin{figure*}[t]
  \centering
  \includegraphics[width=0.6\linewidth]{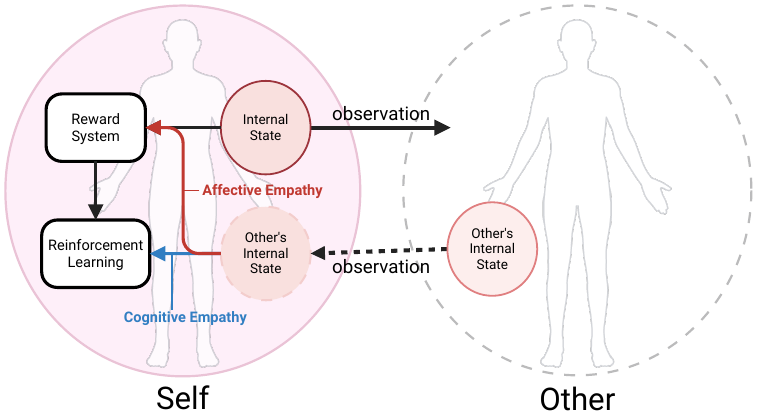}
  \caption{Conceptual diagram of the homeostatic coupling. The homeostatic couplings are illustrated for a single agent (`Self' side).}
  \label{fig:coupling}
     \vspace{-3mm}
\end{figure*}

\vspace{-2mm}
\subsubsection{Homeostatic Couplings}
Therefore, using this toy environment, we explore conditions under which the Possessor will share food with the Partner and prosocially maintain both agents’ energy states at High. We compare the following four conditions in numerical simulations as illustrated in Figure \ref{fig:coupling}: i) The Possessor optimizes only for its own homeostasis ({\bf none} condition). ii) The Possessor can observe the energy state of the Partner but is not specifically motivated to maintain the Partner's homeostasic state ({\bf cognitive} empathy condition). iii) The Possessor does not explicitly observe the energy state of the Partner, but has its own energy state coupled to the Partner’s energy state with a weighting factor ({\bf affective} empathy condition). Specifically, the weighting factor $w=0.5$ and the drive of the Possessor is given by 
\begin{eqnarray}
D = D_{\rm possessor} + w D_{\rm partner}.
\end{eqnarray}
 iv) The final situation combines both cognitive and affective empathy. The Possessor can explicitly observe the energy state of the Partner, and the Possessor’s energy state is also coupled to that of the Partner ({\bf full} empathy condition).

\vspace{-2mm}
\subsection{Extension to Dynamic Environments}
\label{sec:continuous_env}

\begin{figure*}[t]
  \centering
  \includegraphics[width=0.7\linewidth,bb=0 0 1220 471]{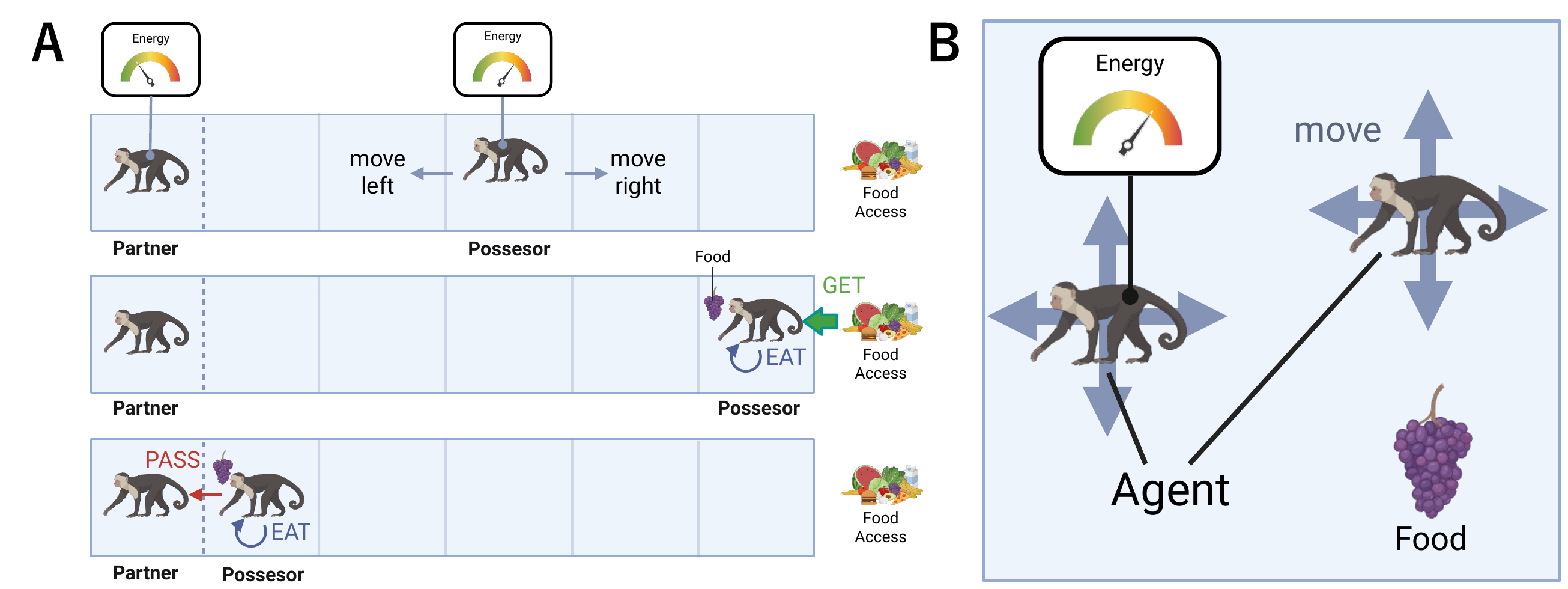}
  \caption{
Overview of the mobile agent environments. {\bf A}: Grid environment. {\bf B}: 2-D continuous field environment.
  }
  \label{fig:grid_trap_env}
       \vspace{-3mm}
\end{figure*}

Next, we investigated the generalization of the findings from the food-sharing environment to 1-D and 2-D environments with mobile agents. The first is a linear grid environment (Figure \ref{fig:grid_trap_env}A), in which the Partner is, once again, trapped on the left side of the grid without access to food. The second mobile environment is one in which both agents can move on a two-dimensional field  (Figure \ref{fig:grid_trap_env}B).

\vspace{-2mm}
\subsubsection{Grid Environment}
\label{sec:append_grid}

In the grid environment (Figure \ref{fig:grid_trap_env}A), the Partner is trapped in the left side of the grid, just as in the food-sharing environment. The Possessor can move left or right and knows its position (there are five positions in the environment). If the Possessor is at the right-most position, next to the food source, it can acquire food by selecting the GET action. Once food is acquired, the Possessor transitions from the without-food state to the with-food state. In the with-food state, the Possessor can select the EAT action and transition itself to a higher energy state.  Alternatively, the Possessor can move and carry the food to the left-most position, at which point the Possessor can choose the PASS action and transition the partner to a higher energy state (and transition itself to the without-food state). The Possessor can choose between five actions: LEFT, RIGHT, EAT, GET, and PASS. Invalid actions (e.g. GET without being in the right-most position) will advance a timestep but are otherwise ignored.

In addition, the energy state of each agent is represented as a continuous variable. The dynamics of the energy state $s^i$ is represented by 
\begin{eqnarray}
s^i_{t+1} = s^i_t - \delta_\text{out} + \delta_\text{in} \cdot I_t.
\end{eqnarray}
In this case, $\delta_\text{out} = 0.003$ and $\delta_\text{in}= 0.1$  is a fixed constant that represents a certain amount of energy consumption and inlet, respectively. $I_t$ is a function that returns 1 when the agent has ingested food, and 0 otherwise. 

In this experiment, the homeostatic setpoint of energy level was 0, and therefore the drive function was given by the squared error $D=\|s^i \|^2$. Agents were trained with a reward scale $\beta=100$. If the energy state of either agent deviated from the range $[-1, 1]$, the episode was terminated.

\vspace{-2mm}
\subsubsection{Continuous 2D--Field Environment}
\label{sec:continuous_env}

\begin{figure*}[t]
  \centering
  \includegraphics[width=0.7\linewidth]{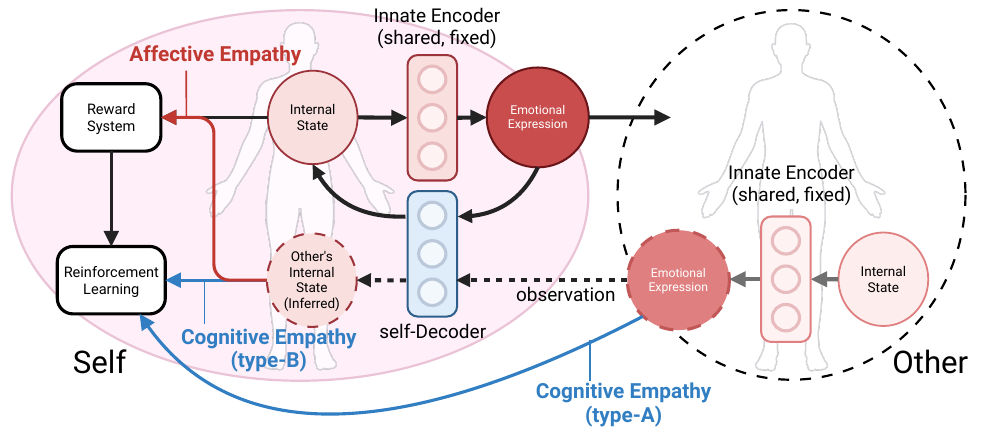}
  \caption{
The innate encoder and learned self-decoder mechanism for autonomous learning of homeostatic coupling. An internal homeostatic state is transformed into an outward emotional expression through a nonlinear encoder function. This function is fixed and identical across agents. An agent can also learn the inverse mapping, decoding emotional expressions back to their underlying homeostatic states, by (self-supervised) training the self-decoder on its own internal states. Once trained, the agent can decode another agent's emotional displays to infer their inaccessible internal states. These inferred states can then be used for cognitive or affective empathy. 
  }
  \label{fig:decoder_settings}
       \vspace{-3mm}
\end{figure*}

In this environment, the agents can move around in a two-dimensional continuous space (Figure \ref{fig:grid_trap_env}B). Now there is no distinction between Partners and Possessors, as both agents can freely move (UP, DOWN, LEFT, RIGHT), GET, EAT, and PASS food. When agents are close enough to each other, the PASS action becomes valid.

The energy state dynamics are carried over from the grid environment (baseline energy expenditure of $\delta_\text{out} = 0.001$ and food energy value of $\delta_\text{in}= 0.1$. Additionally, if an agent's energy level becomes less than -0.7, it is considered to be damaged and cannot move. Therefore, in such a situation, the agent needs to be fed by its partner in order to be able to move again. Finally, the agents have some small risk of becoming injured at each timestep. If both agents are able to move, each agent has a small probability $p=0.0005$ of having an accident: its energy value drops to -$0.7$ and it becomes immobile. Episodes are terminated if either agent's energy value falls below -1. 


\vspace{-2mm}
\subsection{Learning to Infer the Internal States of Others}
\vspace{-1mm}
In the preceding experiments, we assumed that an agent would have direct access to the internal state of another agent. It could either perceive or couple itself to the partner's true internal state, directly provided by the simulation engine. However, in more realistic scenarios – such as social animals in the natural world – states are only partially observable. One must learn prosocial behavior by interpreting observable cues and emotional expressions to infer the internal states of one's partners. 
Among social animals, internal feeling states often manifest externally, for instance in facial emotional expressions, and provide cues to infer the internal generating states \citep{dolensek2020facial,tlaie2024inferring}. Building on this observation, we explore the possibility that agents could predict the homeostatic states of others via the assumption of shared emotion generating mechanisms (physiological response mappings). An agent learns to invert its own mapping between homeostatic states and emotional expressions, and applies it to the expressions of its partners.
To simulate the generation and interpretation of emotional expressions, each agent (`Self' and `Other') contains a fixed encoder that maps internal states to externally observable emotional expressions. The encoder is complemented by a self-decoder that learns to reconstruct one's internal state, given the associated emotional expression state (Figure \ref{fig:decoder_settings}) . The fixed encoders are identical across agents, reflecting the assumption that animals of the same species are governed by similar mechanisms for generating spontaneous emotional expressions. It then becomes possible to learn prosocial behaviors responsive to the needs of others.
\vspace{-2mm}
\subsubsection{Self- and Other-Decoding Experiments}
We conducted experiments on homeostatic inference using the food-sharing and grid environments. The Possessor agent attempted to learn prosocial behavior by first learning its own internal dynamics through self-supervision. Emotional expressions were modeled as publicly observable 5-dimensional continuous vectors $\vv_c$, generated from internal states $s^i_c$ via an MLP with a 20-dimensional hidden layer and ReLU activation. The weights of this generator network were initialized randomly, fixed, and shared across both agents. The Possessor learned a self-decoder $\text{dec}_\theta(\cdot)$ with parameters $\theta$, trained using data collected during PPO training to minimize the mean-squared reconstruction error
\begin{eqnarray}
J(\theta) = \frac{1}{N}\sum_{n=1}^N \| \text{dec}_\theta(\vv_{\text{possessor},n}) - s^i_{\text{possessor},n}\|^2,
\end{eqnarray}
where $N$ represents the mini-batch size, and $\vv_\text{possessor}$ is the Possessor's emotional expression. The decoder network was a mirror image of the generator network, with a 5-dimensional input, 20-dimensional hidden layer, and ReLU activation.
We compared three empathy conditions: cognitive, affective, and full. In each case, the Possessor decoded the Partner’s emotional expression and inferred the Partner’s internal state ${\tilde s}^i_\text{partner}= \text{dec}_\theta(\vv_{\text{partner}})$. Based on that inferred internal state ${\tilde s}^i_\text{partner}$, the partner's drive ${\tilde D}_\text{partner}=\|{\tilde s}^i_\text{partner}\|^2$ is estimated. Under affective empathy, the Possessor's new drive state is calculated as a weighted sum of its own drive and its partner's estimated drive $D = D_{\rm possessor} + w {\tilde D}_{\rm partner}$. Again, we used $w=0.5$. 

For cognitive empathy, the Partner's emotional expression was used in two different ways. The emotion vector $\vv_\text{partner}$ could serve as a direct observation input to RL (cognitive empathy type-A in Figure \ref{fig:decoder_settings}), or it could have been further decoded by the self-decoder into Partner's inferred state ${\tilde s}^i_\text{partner}$ before being input as an observation to RL (cognitive empathy type-B). To examine whether training the self-decoder influences prosocial behavior, we also conducted ablation experiments in which the decoder was not trained under either the type-A or type-B condition. Each condition was tested 20 times, and the average episode length along with its 95\% confidence interval was calculated.
\begin{figure}[t]
  \centering
  \includegraphics[width=\linewidth]{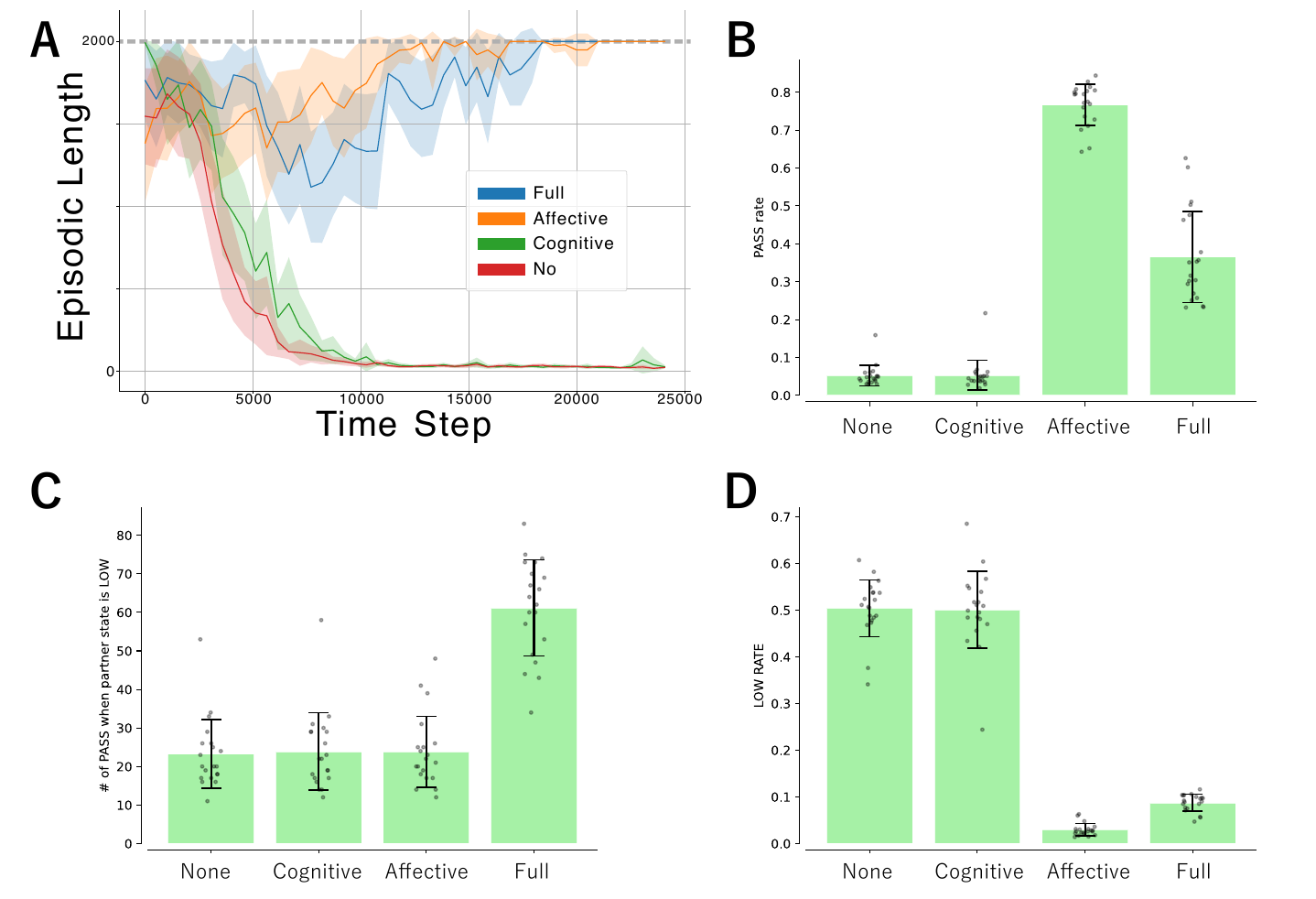}
  \vspace{-5mm}
\caption{
Learning and behavior evaluation in a food sharing environment. {\bf A}: Learning curves with performance measured by episode duration with both agents alive ({\emph n}=20, 95\% confidence intervals). Only the conditions that implement affective empathy (affective and full) result in long episode durations. {\bf B}: PASS behavior selection rate out of 1,000 steps of the test run. Possessor agents in the affective condition learn to frequently pass food to the Partner. {\bf C}: Count of PASS actions conditioned on the Partner being in the Low energy state. Possessor agents in the full empathy condition learn to selectively pass food to the Partner when it is most needed. {\bf D}: Likelihood of the Partner agent to be in the LOW state. 
  }
  \label{fig:fs_results}
       \vspace{-3mm}
\end{figure}

\begin{figure}[t]
  \centering
  \includegraphics[width=\linewidth]{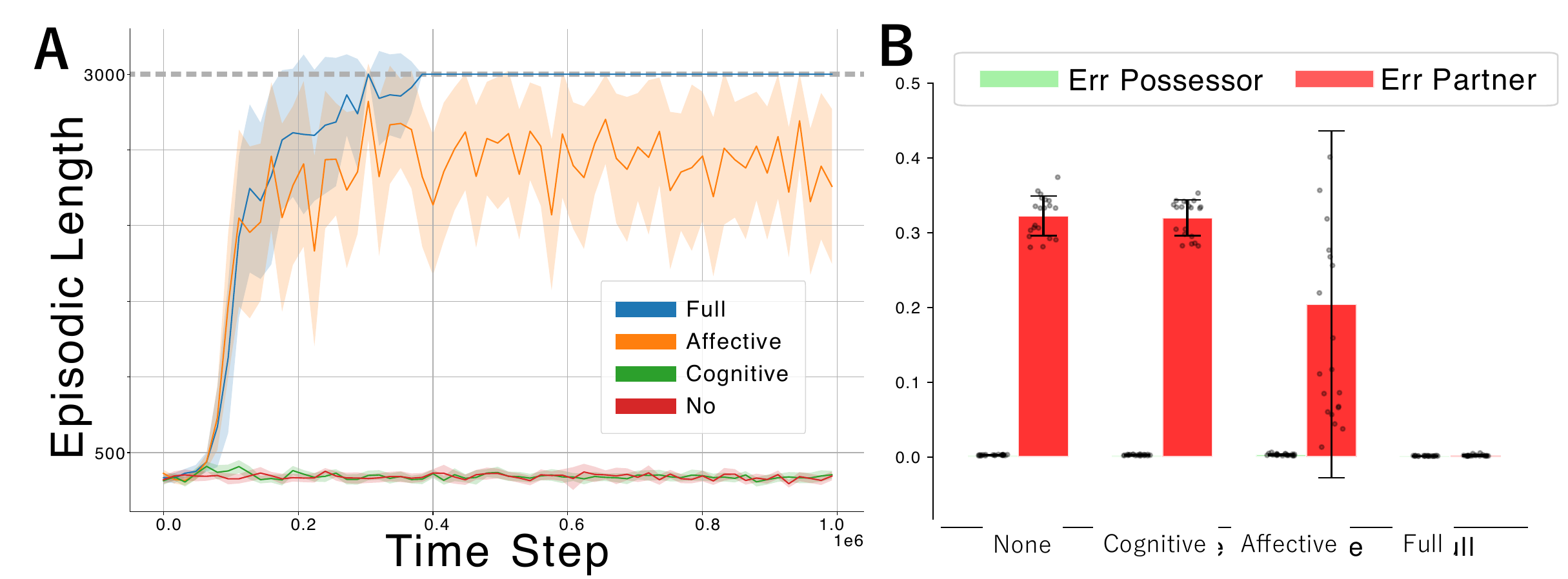}
  \vspace{-5mm}
\caption{
Performance in the linear grid mobile environment. {\bf A}: Learning curves with performance measured by episode duration with both agents alive ({\emph n}=20). {\bf B}: Homeostatic drives of agents ($D_{\rm possessor}$ and $D_{\rm partner}$) averaged over 1,000 time steps.
  }
  \label{fig:grid_env}
\end{figure}

\begin{figure}[t]
  \centering
  \includegraphics[width=\linewidth]{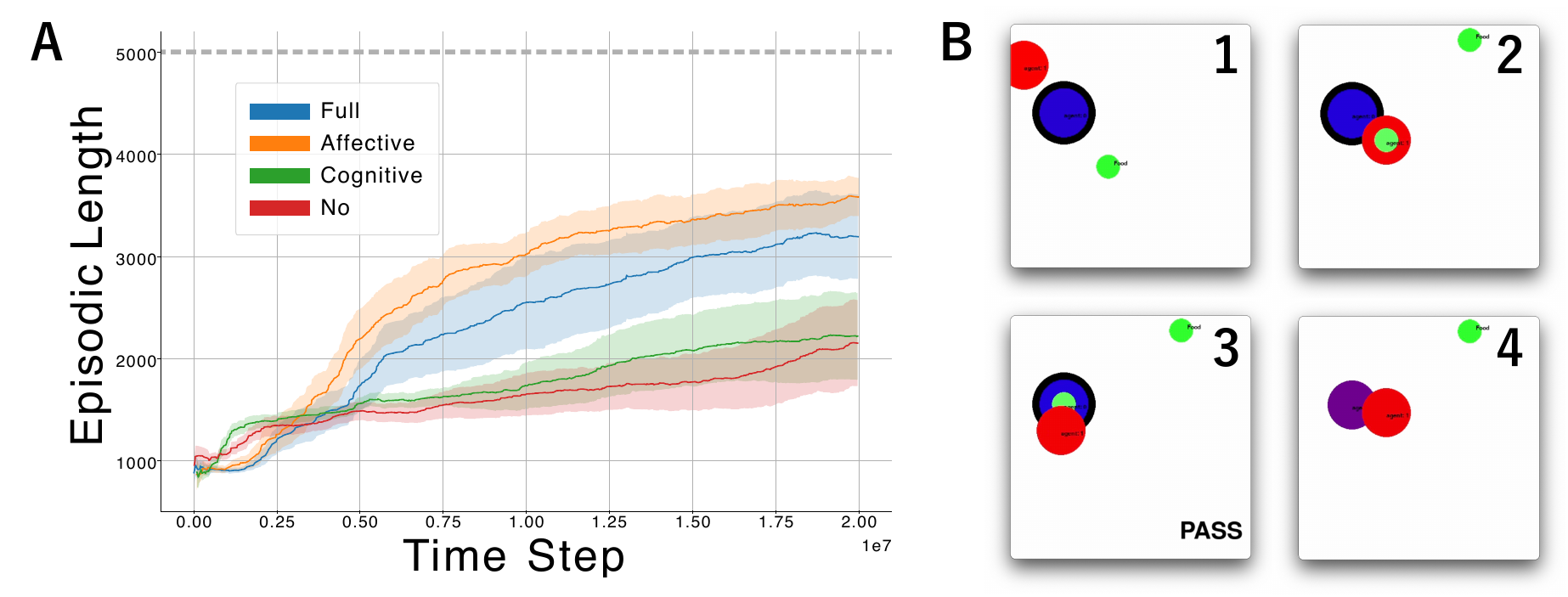}
   \vspace{-5mm}
 \caption{
Performance in the 2-D field mobile environment. {\bf A}: Learning curves with performance measured by episode duration with both agents alive ({\emph n}=20) and 95\% confidence interval.  {\bf B}: An example of helping behavior observed in the Affective condition. The action sequence progresses in order of the numbers in the top right corner of each panel. The blue agent is immobilized due to its low energy level. The red agent collects a green food pellet and returns to share it with the blue agent, turning it purple (replenishing some energy) and restoring it to mobility.  
  }
  \label{fig:trap_env}
       \vspace{-3mm}
\end{figure}

\vspace{-2mm}
\section{Results}
\vspace{-1mm}
\subsection{Food Sharing Environment}
 
Average learning curves are shown in Figure \ref{fig:fs_results}A. Performance is evaluated by episode duration, with a maximum length of 2,000 steps. In the none condition, the PASS action is rarely selected (Fig. \ref{fig:fs_results}B) and episode lengths did not increase over training. Similar results are obtained in the cognitive empathy condition, in which the Possessor observes, but is not motivated by, the Partner's homeostatic state. On the other hand, the homeostatic states of both agents are maintained under the affective and full empathy conditions, leading to long episode durations. In the affective condition, the Possessor does not have explicit knowledge of the Partner’s energy state and so frequently chooses the PASS action to help the Partner maintain homeostasis, thereby also regulating its own homeostatic state because it is coupled to that of the Partner’s. This suggests that a strategy was acquired to maintain homeostasis between the two by supplying an excess of food to the Partner.

Figure \ref{fig:fs_results}C-D supports this speculation. Considering only the times when the Partner is in a Low energy state, Full condition agents selected the PASS action more often than Affective condition agents (Figure \ref{fig:fs_results}C). This implies that the Possessor in the full condition learned to be sensitive to the Partner's internal state, more frequently selecting the PASS action when the Partner's state was low. The PASS frequency for the affective condition was similar to the other conditions, but this is likely because there are few opportunities for the Partner's state to become LOW in the affective condition (see Figure \ref{fig:foodshare_env}D). Altogether, these results suggest that a minimal requirement for prosocial behavior is an internalized motivation for the well-being of others.

\vspace{-2mm}
\subsection{Mobile Environments}
Figures \ref{fig:grid_env} and \ref{fig:trap_env} show the results of optimization in each mobile environment. No prosocial behavior was learned in the none and cognitive conditions, and episode durations remained short (Figure \ref{fig:grid_env}A and \ref{fig:trap_env}A). 

As in Figure \ref{fig:grid_env}B, the variance of the homeostatic drive of the Partner ($D_{\rm partner}$) is large in the affective condition. One possible explanation is that the Possessor agent cannot observe the energy state of its Partner, therefore the Partner agent is fed indiscriminately in the affective condition, at various energy values (Appendix B, Figure \ref{fig:grid_dist}). Figure \ref{fig:trap_env}B captures an example sequence of prosocial behavior observed in 2-D field environment with the affective condition. 
\begin{figure}[!t]
  \centering
  \includegraphics[width=0.7\linewidth]{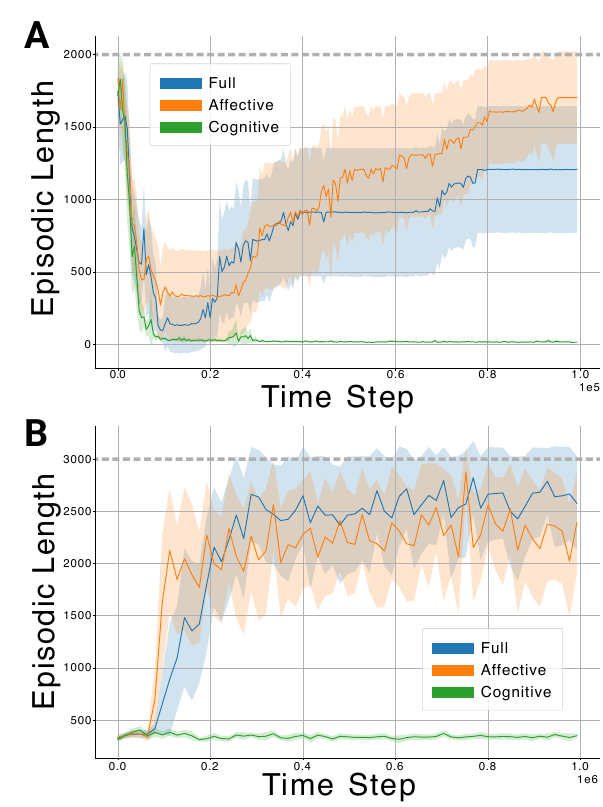}
  \vspace{-2mm}
  \caption{
  Results of self-decoder experiment. {\bf A}: Food-sharing environment. {\bf B}: Grid environment. 
  }
  \label{fig:decoder_res}
       \vspace{-3mm}
\end{figure}

\vspace{-6mm}
\subsection{Inferring the Internal States of Others}
Figure \ref{fig:decoder_res} show the mean and 95\% confidence interval over 20 trials in the food-sharing (panel A) and grid (panel B) environments. Consistent with preceding experiments, prosocial behavior was not acquired under the cognitive condition. Under the full and affective conditions, by contrast, the Possessor successfully learned prosocial behavior. It learned to use its self-decoder to decode the states of the Partner, and then to empathize accurately. These results suggest that autonomous acquisition of empathetic behavior is possible through learning the mapping between internal bodily states and one’s own emotional expressions.

Table \ref{fig:decoder_res} presents the results of a comparison of final test performance (group longevity) after learning in the grid environment, under various conditions: when cognitive empathy was based on directly perceived expressions (type-A) or decoded expressions (type-B), and when the self-decoder was either trained or not. In general, under the cognitive condition, prosocial behavior was not acquired in any case, whereas in the other conditions, training of self-decoder promoted the emergence of prosocial behavior. In the full condition, prosocial behavior was successfully acquired in both type-A and type-B without significant difference. 

Unexpectedly, in the grid environment, even an untrained, randomly initialized self-decoder led to extended survival of the Partner agent, albeit with lower performance. Since such a phenomenon was not observed in similar comparisons in the food-sharing environment, it is possible that this result arises from the internal state being represented as continuous variables. A deeper understanding of this spontaneous acquisition of prosocial behavior will require further experiments and analyses. 

\begin{table}[t]
    \centering
    \begin{tabular}{cccc}
        \toprule
                               & cognitive       & affective$^\dagger$\tnote{$\dagger$} & full \\
        \midrule
        A w/ train & $356 \pm 29$  & \multirow{2}{*}{$\bf 2147 \pm 558$} & $ 2625 \pm 499$\\
        B w/ train & $342 \pm 36$ &                                                      & $\bf  2812 \pm 368$ \\ \midrule
        A w/o train & $368 \pm 31$ & \multirow{2}{*}{$1628 \pm 795$} & $1125 \pm 802$ \\
        B w/o train & $295 \pm 26$ &                                                    & $1182 \pm 791$ \\
        \bottomrule
    \end{tabular}
    \begin{tablenotes}[para,online,normal] 
	\item[$\dagger$]\scriptsize{The affective only condition does not have the cognitive empathy.}
    \end{tablenotes}
         \vspace{-3mm}
    \caption{
    Ablation studies in the grid environment.
    }
    \label{tab:ablation}
         \vspace{-3mm}
\end{table}

\section{Discussions and Concluding Remarks}
\vspace{-1mm}
This study investigated the emergence of prosocial behavior in simple RL agents motivated by homeostatic self-regulation. We found that prosocial behavior (food sharing, leading to extended mutual well-being) only occurred reliably under affective empathy, when the homeostatic states of agents were coupled together. Perception of a partner’s state of need did not, on its own, drive prosocial behavior. The combination of cognitive and affective empathy in the full condition drove more sensitive sharing behavior.  

One possible future direction is to characterize prosocial behaviors using more information-theoretic metrics. It may be possible to achieve and maintain high group well-being by implementing a form of mutual information \citep{jaques2019social} in sequential social dilemmas \citep{leibo2017multi}, such that successfully self-regulating agents can gain influence and induce the well-being of others. 

As demonstrated in our final experiment, it is possible to infer the internal states of others using a self-decoder model learned from one's own internal dynamics. This process would better resemble the mirror neuron system, hypothesized to support emotion recognition and empathic behavior in humans and other animals \citep{rizzolatti2005mirror,iacoboni2009imitation}. For example, neurons in the inferior parietal lobule activate both during the observation and imitation of emotions; they can then trigger activity through the insula into the limbic system, known to activate during the firsthand experience of emotional feelings \citep{carr2003neural}. 
Simulations of mirror neurons have also been conducted in the context of cognitive developmental robotics \cite{asada2009cognitive}, where models have been proposed that can predict the visual sequence resulting from one's own actions based on observed image sequences of others' actions, using predictors trained on first-person visual experiences \cite{seker2022imitation}. In this context, our results suggest that a self-decoder model of interoceptive systems is a crucial factor in the acquisition of prosocial behavior.

\appendix

\section{Appendix A: Tables of Hyperparameters}
 The table shows the network and PPO hyper--parameters for each experiment (Table \ref{tb:fs}--\ref{tb:2d}). The food flag, the movable flag, and the positional information in Grid environment are encoded by the one-hot representation.
 
\begin{table}[H]
\caption{Hyper-parameters of Food Sharing Environment}
\begin{center}
\begin{tabular}{c|c}
Exteroception dim & none  \\
Interoception dim & 1 (energy state)  \\
Interoception dim of other agent &  1 (energy state)\\ (cognitive empathy) &  \\
Hidden dim &  16\\
LSTM hidden state dim  &  16 \\
Total time steps & 25,000\\
Learning rate & 0.001 \\
\# of samplers & 16\\
Sampling steps & 32\\
Discount factor ($\gamma$)& 0.99 \\
GAE lambda & 0.95\\
Number of minibatches & 4\\
Update epochs & 4\\
Normalizing advantage & True\\
Clip coefficient of policy update & 0.1 \\
Value clipping loss & True \\
Entropy coefficient & 0.01\\
Value loss coefficient & 0.5\\
Maximum gradient norm & 0.5\\
\end{tabular}
\end{center}
\label{tb:fs}
\end{table}

\begin{table}[H]
\caption{Hyper-parameters of Grid Environment}
\begin{center}
\begin{tabular}{c|c}
Exteroception dim & 5 (position)\\ & + 2 (food flag)  \\
Interoception dim & 1 (energy state)  \\
Interoception dim of other agent & 1 (energy state)\\ (cognitive empathy) &  \\
Hidden dim&  32\\
LSTM hidden state dim  & 32 \\
Total time steps & 1,000,000\\
Learning rate & 0.001 \\
\# of samplers & 16\\
Sampling steps & 100\\
Discount factor ($\gamma$)& 0.99 \\
GAE lambda & 0.95\\
Number of minibatches & 4\\
Update epochs & 4\\
Normalizing advantage & True\\
Clip coefficient of policy update & 0.1 \\
Value clipping loss & True \\
Entropy coefficient & 0.01\\
Value loss coefficient & 0.5\\
Maximum gradient norm & 0.5\\
\end{tabular}
\end{center}
\label{tb:grid}
\end{table}

\begin{table}[!h]
\caption{Hyper-parameters of 2D Field Environment}
\begin{center}
\begin{tabular}{c|c}
Exteroception dim & 2 (position)\\ &  + 2 (food position) \\
& + 2 (food flag)\\ & + 2 (movable flag)\\
Interoception dim & 1 (energy state)  \\
Interoception dim of other agent & 1 (energy state)\\ (cognitive empathy) &   \\
Hidden dim&  64\\
LSTM hidden state dim  &  64 \\
Total time steps & 20,000,000\\
Learning rate & 0.001 \\
\# of samplers & 16\\
Sampling steps & 1024\\
Discount factor ($\gamma$)& 0.99 \\
GAE lambda & 0.95\\
Number of minibatches & 2\\
Update epochs & 4\\
Normalizing advantage & True\\
Clip coefficient of policy update & 0.1 \\
Value clipping loss & True \\
Entropy coefficient & 0.0\\
Value loss coefficient & 0.3\\
Maximum gradient norm & 0.5\\
\end{tabular}
\end{center}
\label{tb:2d}
\end{table}

\section{Appendix B: Histograms of the Energy States}
\begin{figure}[H]
  \centering
  \includegraphics[width=\linewidth,bb=0 0 659 479]{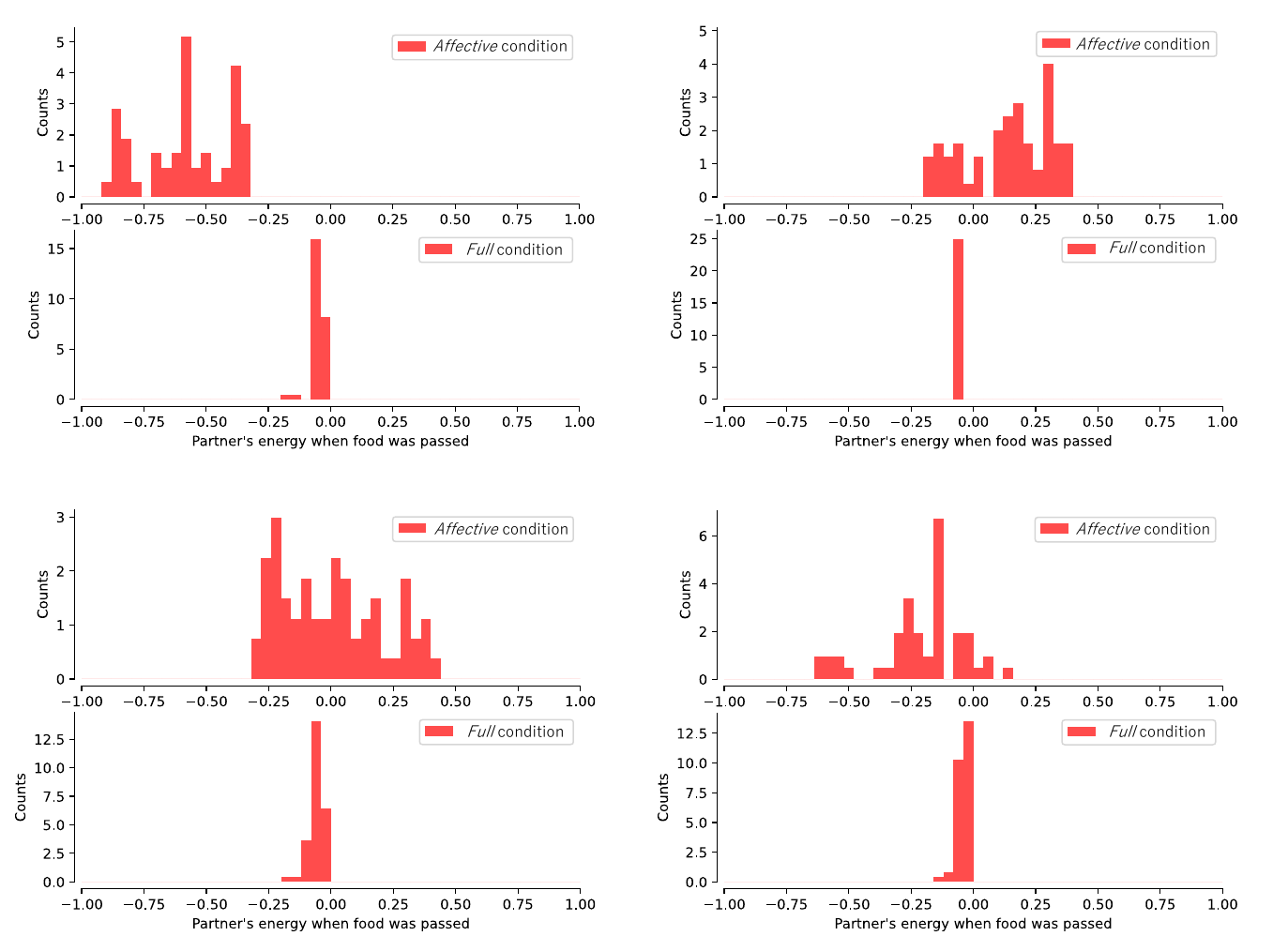}
  \caption{
  Typical histograms of the energy state of the Partner agent when it ingests food during the 2,000-step test run after optimization. 
  }
  \label{fig:grid_dist}
\end{figure}

\section*{Acknowledgements}
We thank Boris Gutkin for discussion of our early version of the paper. This work was supported by a grant from the Foresight Institute to KM, and Japan Society for the Promotion of Science KAKENHI grant 24K23892 to NY. Figures were created using BioRender.com.

\footnotesize
\bibliographystyle{apalike}
\bibliography{reference} 

\end{document}

%% file: math_commands.tex
\usepackage{amsmath,amsfonts,bm}

\def\eqref#1{equation~\ref{#1}}

\def\1{\bm{1}}

\def\vv{{\bm{v}}}

\DeclareMathAlphabet{\mathsfit}{\encodingdefault}{\sfdefault}{m}{sl}
\SetMathAlphabet{\mathsfit}{bold}{\encodingdefault}{\sfdefault}{bx}{n}